\documentclass[letterpaper, 10 pt, conference]{ieeeconf}  

\IEEEoverridecommandlockouts                              

\usepackage{url}
\usepackage{cite}
\usepackage{balance}
\usepackage{graphicx}
\usepackage{amsfonts}
\usepackage{fancyhdr}
\usepackage{comment}
\usepackage{times}
\usepackage{amsmath}
\usepackage{changepage}
\usepackage{amssymb}
\usepackage{enumerate} 
\usepackage{algorithmic}
\usepackage{bm}
\usepackage{calligra}
\usepackage{multirow}
\usepackage{tabularx}

\usepackage{booktabs}
\usepackage{mathtools}
\usepackage{arydshln}
\usepackage{latexsym} 
\usepackage{amsmath}
\usepackage{amssymb}
\usepackage{color}
\usepackage[ruled,vlined]{algorithm2e}
\usepackage{float}
\usepackage{tabstackengine}
\usepackage{siunitx}
\usepackage{colortbl}
\usepackage{hhline}
\usepackage{bookmark}
\usepackage{dirtree}
\usepackage{subfigure}

\def\BibTeX{{\rm B\kern-.05em{\sc i\kern-.025em b}\kern-.08em
    T\kern-.1667em\lower.7ex\hbox{E}\kern-.125emX}}
\usepackage[font=small,labelfont=bf,tableposition=top]{caption}
\usepackage{blindtext}

\usepackage{fancyhdr}

\begin{document}

\fancypagestyle{withfooter}{
  \renewcommand{\headrulewidth}{0pt}
  \fancyfoot[C]{\footnotesize Accepted to the IEEE ICRA Workshop on Field Robotics 2024}
}

\title{\LARGE \bf
Are We Ready for Planetary Exploration Robots? 
\\ The TAIL-Plus Dataset for SLAM in Granular Environments
}

\author{Zirui Wang$^{1}$$^{*}$, Chen Yao$^{1}$$^{*}$, Yangtao Ge$^{1}$$^{*}$, Guowei Shi$^{1}$$^{*}$, Ningbo Yang$^{1}$, Zheng Zhu$^{1}$, \\Kewei Dong$^{1}$, Hexiang Wei$^{2}$, Zhenzhong Jia$^{1}$, Jing Wu$^{1}$
\thanks{$^{1}$Department of Mechanical and Energy Engineering, Southern University of Science and Technology (SUSTech), Shenzhen, 518055, China.}
\thanks{$^{2}$The Hong Kong University of Science and Technology (Guangzhou), Nansha, Guangzhou, 511400, Guangdong, China.}
\thanks{*Equal contribution.}
\thanks{This work was supported in part by National Science Foundation of China (62203205 \& U1913603). The views, opinions, findings and conclusions
 reflected in this publication are solely those of the authors and
 do not represent the official policy or position of the NFSC.}
}

\let\oldtwocolumn\twocolumn
\renewcommand\twocolumn[1][]{%
    \oldtwocolumn[{#1}{
        \begin{center}
            \vspace{1mm}
           \includegraphics[width=0.99\textwidth]{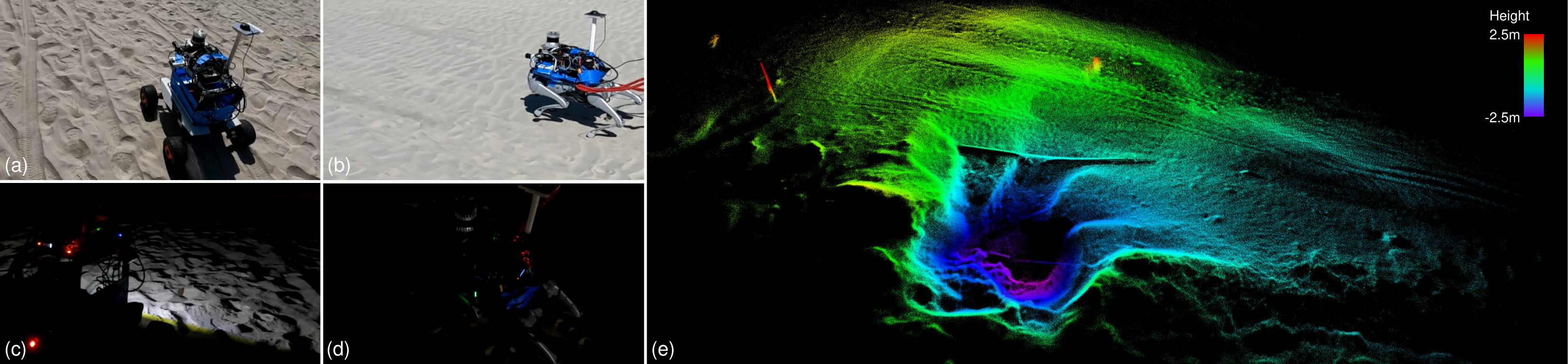}
           \captionof{figure}{Filed experiments in diverse sandy terrains utilizing both wheeled (a) and quadruped (b) robot. These two robot platforms carried the same sensor suite. The headlights of the wheeled robot were turned on at night during data collection (c) while there was no additional lighting for the quadruped robot (d). (e) shows the 3D pointcloud map of the terrain traversed by the robot shown in (a). The pointcloud map is generated from 3D LiDAR.}
           \label{fig:intro_picture}
           \vspace{2.8mm}
        \end{center}
    }]
}

\maketitle

\thispagestyle{withfooter}
\pagestyle{withfooter}

\begin{abstract}
So far, planetary surface exploration depends on various mobile robot platforms. The autonomous navigation and decision-making of these mobile robots in complex terrains largely rely on their terrain-aware perception, localization and mapping capabilities. 
In this paper we release the TAIL-Plus dataset, a new challenging dataset in deformable granular environments for planetary exploration robots, which is an extension to our previous work, TAIL (Terrain-Aware multI-modaL) dataset. 
We conducted field experiments on beaches that are considered as planetary surface analog environments for diverse sandy terrains.
In TAIL-Plus dataset, we provide more sequences with multiple loops and expand the scene from day to night. 
Benefit from our sensor suite with modular design, we use both wheeled and quadruped robots for data collection. The sensors include a 3D LiDAR, three downward RGB-D cameras, a pair of global-shutter color cameras that can be used as a forward-looking stereo camera, an RTK-GPS device and an extra IMU.
Our datasets are intended to help researchers developing multi-sensor simultaneous localization and mapping (SLAM) algorithms for robots in unstructured, deformable granular terrains.
Our datasets and supplementary materials will be available at \url{https://tailrobot.github.io/}. 
\end{abstract}

\section{Introduction}
\label{sec:introduction}

With the development of robotic techniques, mobile robots have already achieved huge success in the planetary surface exploration missions\cite{ding_surface_2022,layered_2022,ground_2023}. 
Due to the lack of prior knowledge of planetary surface and the unstructured deformable terrains, planetary exploration robots could be exposed to potential risks, such as sinkage and slippage, which challenge the robot's mobility and locomotion. 
Under the constraints of high communication latency, high packet loss rate, low bandwidth and periodic relay communication, it is impossible to achieve real-time teleoperation or get real-time feedback data from robots.
These challenges require planetary exploration robots to have a certain level of autonomy\cite{telezhurong,Perseverance,perstraverse}. The autonomous navigation and reliable decision-making of these robots in complex terrains heavily rely on their localization and terrain perception capabilities.
In recent years, the development of multi-sensor simultaneous localization and mapping (SLAM) solutions \cite{shan2021lvisam,zheng2022fastlivo,r2live_2021,r3live_2022} shows the immense potential of multi-sensor fusion in field robot applications. Multi-sensor fusion can combine the advantages of different sensors and ensure the robustness and accuracy localization and environment perception in complex terrains, holding the promise of safe locomotion in the planetary surface exploration missions.

The research on SLAM systems heavily relies on various datasets that can be used for evaluations and repeatable experiments.
For planetary exploration robot applications, researchers prefer selecting deserts or beaches as planetary analog environments because of their unstructured, deformable sandy terrains. The MADMAX dataset \cite{meyer2021madmax} used the Sensor Unit for Planetary Exploration Rovers (SUPER) for data collection in the Moroccan desert. It provided sequences with a total trajectory length of 9.2km. But this work focuses on visual‐inertial navigation systems, and thus did not utilize other kinds of sensors like LiDAR (light detection and ranging). 
The Katwijk rover dataset\cite{hewitt2018katwijk} used the Heavy-duty planetary rover (HDPR) platform to traverse along a beach near Katwijk. The data were obtained from different sensors including ToF camera and 3D LiDAR. However, the sensors were not time-synchronized, which is considered as an important issue in multi-sensor fusion systems. It is noticed that the above previous works (MADMAX and Katwijk) only focus on wheeled robot platforms for analog planetary exploration. \cite{scirobotics.ade9548} conducted planetary analog environments exploration missions achieved by a team of legged robots. The experiments show the high locomotion capability and robustness of quadruped robots to traverse through deformable, unstructured, granular terrains where the mobility of wheeled platforms is limited.   

Considering these, we utilize both wheeled and legged robots to develop datasets for multi-sensor fusion SLAM in unstructured, deformable granular environments. In this paper, we present TAIL-Plus dataset, which is build upon our previous work, \textbf{T}errain-\textbf{A}ware Mult\textbf{I-}Moda\textbf{L} \textbf{(TAIL)} dataset\cite{tail}. 
The original TAIL dataset consists of 14 sequences collected in soft terrains during daytime. In TAIL-Plus dataset, we provide more sequences with multiple loops for testing loop closure methods for long-term SLAM. Moreover, we conducted field experiments at night shown in Fig. \ref{fig:intro_picture}(b) and (c) to extend the scene from day to night for providing visual failure scenarios.

We hope the release of our datasets can help researchers developing multi-sensor localization and mapping algorithms for field robot perception in unstructured granular environments.

\section{System overview}
\label{sec:system}

\begin{figure}[ht!]
\vspace{0.2cm}
\centering
\includegraphics[scale=0.45]{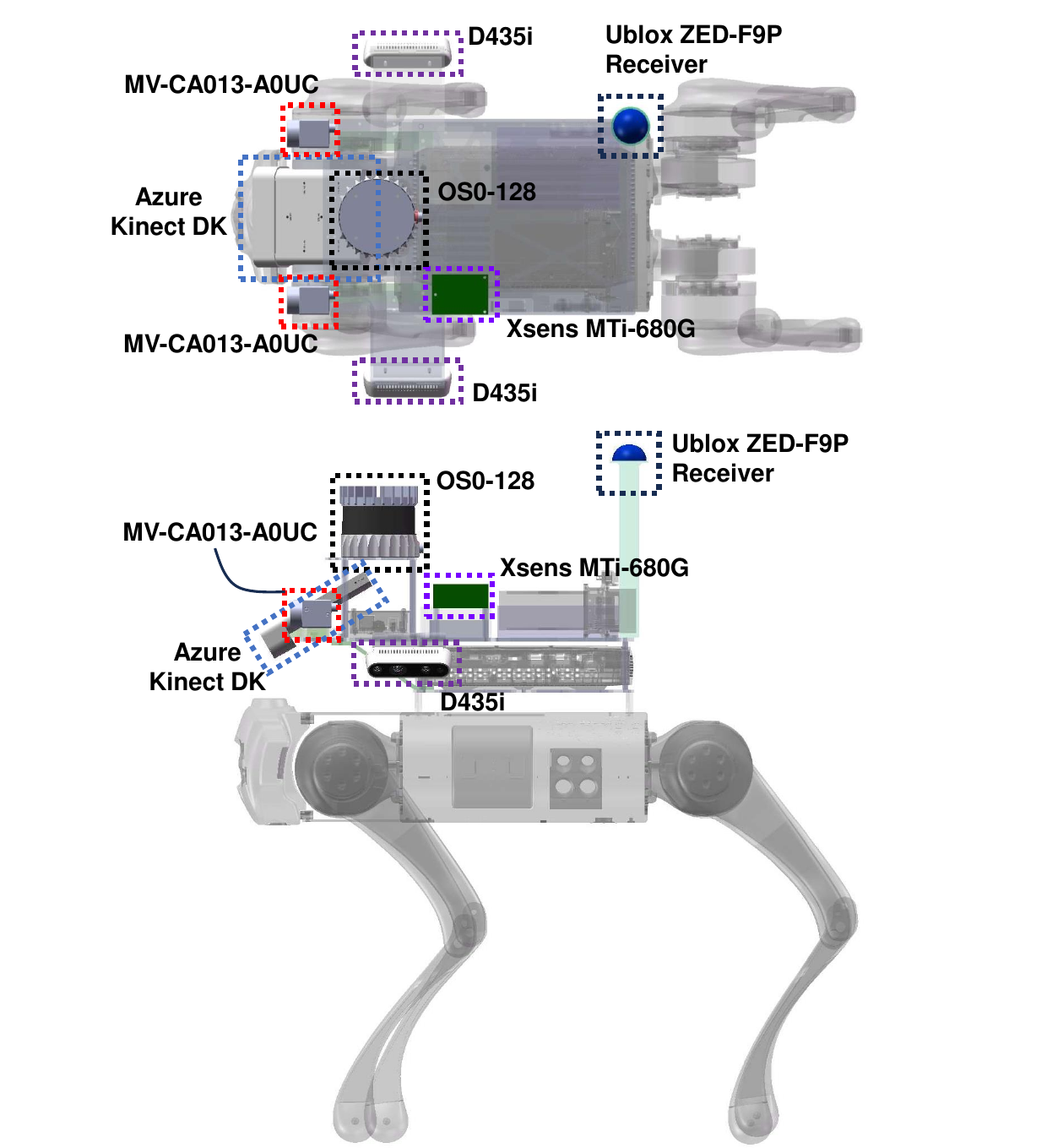}

\vspace{-2mm}
\caption{The sensor suite and the quadruped robot platform. This sensor suite is also installed on wheeled robot, which is shown in Figure \ref{fig:intro_picture}(a).
}
\label{fig:sensor_picture}
\vspace{-0.2cm}
\end{figure}

The wheeled and legged platforms we used are the same as our previous work\cite{tail}.
The sensor suite is shown in Fig. \ref{fig:sensor_picture}. The sensors used for data collection include:
\begin{itemize}
\item 1 $\times$ 3D LiDAR: Ouster OS0-128, 10Hz
\item 3 $\times$ downward RGB-D camera: one Azure Kinect DK (30Hz) and two Realsense D435i cameras (15Hz)
\item 2 $\times$ global-shutter color camera: two HKVision MV-CA013-A0UC cameras with MVL-HF0824M-10MP lenses that can be used as a stereo camera, 30Hz
\item 1 $\times$ RTK-GPS device: Ublox ZED-F9P (GPS: global positioning system; RTK: real-time kinematic)
\item 1 $\times$ extra IMU: Xsens MTi-680G, 100Hz
\item 1 $\times$ built-in robot odometry: wheel odometry of wheeled robot ScoutMini (50Hz), leg odometry of quadruped robot Unitree-GO1 (500Hz)

\end{itemize}

All sensors in the sensor suite shown in Fig. \ref{fig:sensor_picture} are hardware time-synchronized for multi-sensor fusion applications. We use Teensy 4.1, a microcontroller unit, to obtain the PPS (pulse per second) signal form RTK-GPS receiver. Once receiving a PPS signal, the microcontroller forwards preconfigured  trigger signals to different sensors at the required frequencies and intervals. All timestamps of collected data are in the GPS clock.

\begin{figure}[t!]
  \vspace{0cm}
  \centering
  \includegraphics[scale=0.31]{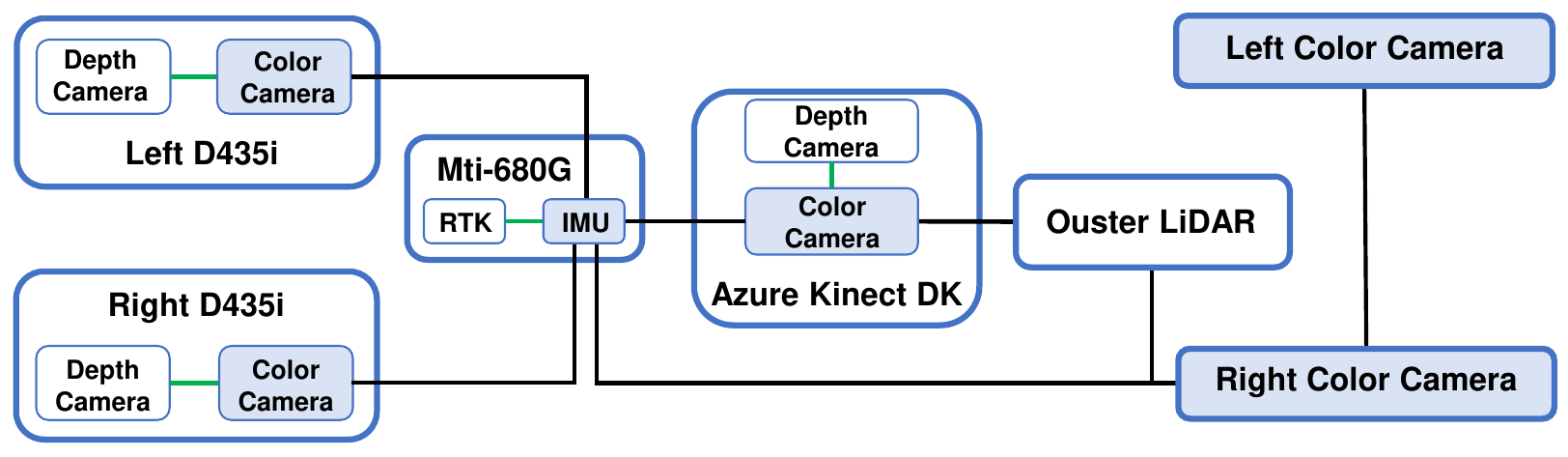}
  \vspace{-0.1cm}
  \caption{The calibration chain. Black lines: the extrinsic parameters are obtained by different calibration methods. Green lines: the extrinsic parameters are provided by factory calibrations or measured from the CAD model.}
  \label{fig:calib}
  \vspace{-0.2cm}
\end{figure}

\begin{figure}[t!]
  \vspace{0cm}
  \centering
  \includegraphics[scale=0.42]{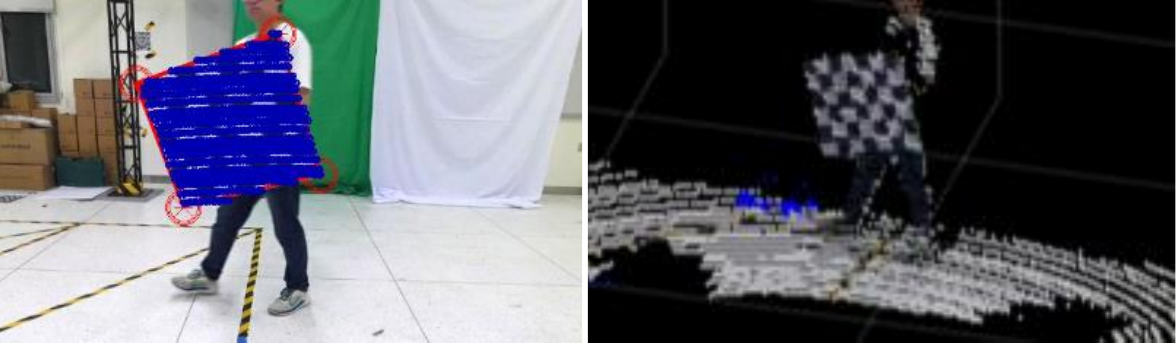}
  \vspace{0cm}
  \caption{The extrinsic calibration between Ouster OS0-128 and Azure Kinect DK color camera. Left: projected checkerboard pointclouds on the color image. Right: the colored pointclouds.}
  \label{fig:calib-cll}
  \vspace{-0.4cm}
\end{figure}

\begin{figure*}[ht!]
  \vspace{0cm}
  \centering
  \includegraphics[scale=0.55]{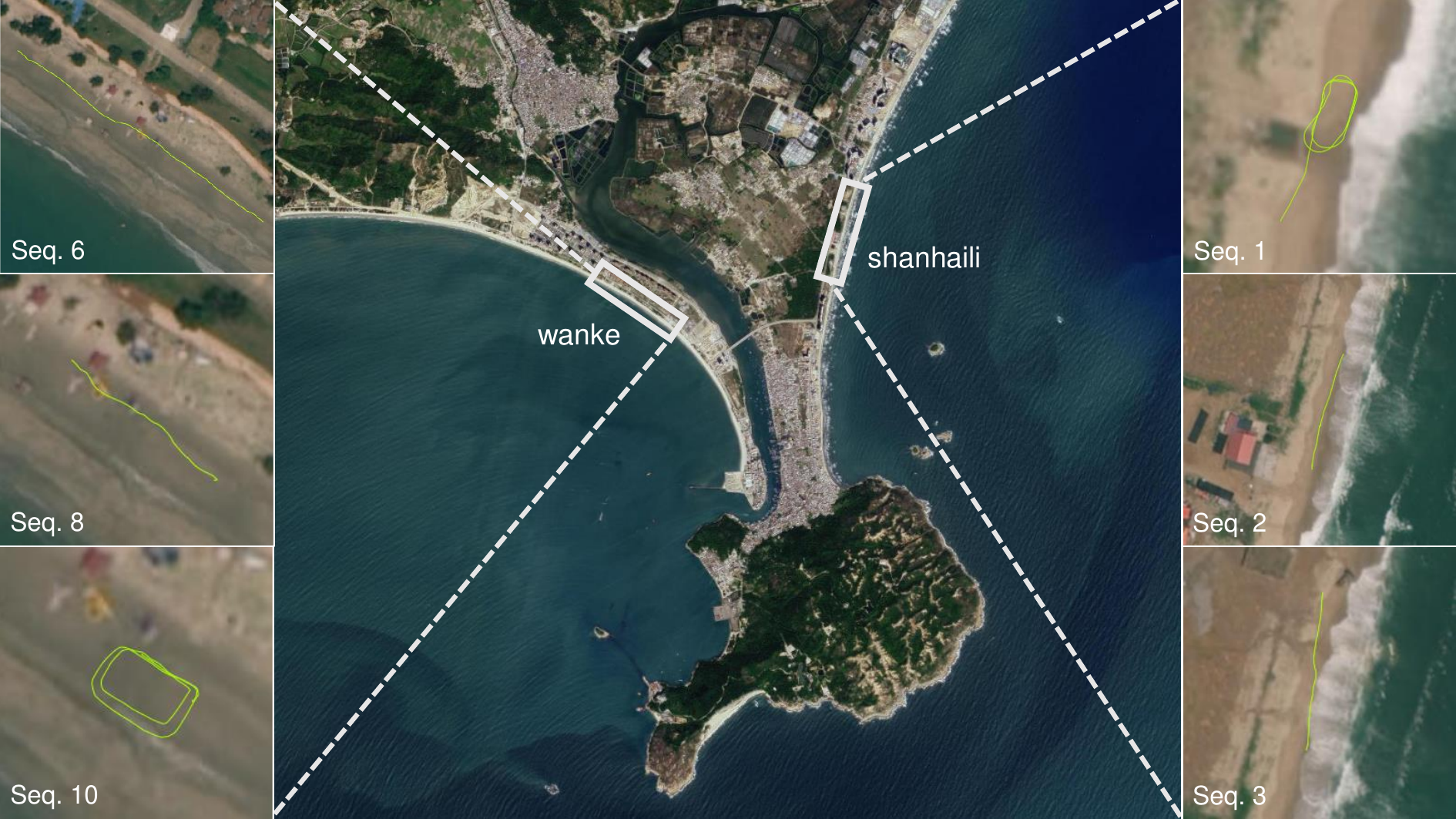}
  \caption{Overview of the field experiment locations and trajectories of part of our TAIL-Plus sequences. The experiment locations are at the Double-Moon Bay, Guangdong Province. Its west coast is called \texttt{wanke} and the east coast is called \texttt{shanhaili}.}
  \label{fig:sample_data}
  \vspace{-0.2cm}
\end{figure*}

\begin{table*}[ht!]
\vspace{0.5cm}
\centering 
\caption{Overview of sequences}
\setlength{\tabcolsep}{2.0mm}{
\resizebox{2.0\columnwidth}{!}{
\begin{tabular}{cllllccccc}
\toprule[0.04cm]
 Datasets & Platforms & Locations & Motions & Sequneces Description & Seq. & T[s] & $||\overline{\bm{v}}||[m/s]$ & Lighting & Appearance \\ \bottomrule[0.03cm]
\multirow{14}{*}
{\begin{tabular}[c]{@{}c@{}}
TAIL\cite{tail}
\end{tabular}} 
&  \multirow{8}{*}
{\begin{tabular}[c]{@{}c@{}}
Wheeled \\ Robot
\end{tabular}} 
&  \multirow{3}{*}{shanhaili} 
 & slow & backward & 1 & 162 & 0.4  & day, weak & grassy, coarse sand \\
 &  &  & medium & forward & 2 & 177 & 0.5  & day, normal & coarse sand \\ 
 &  &  & fast & forward & 3 & 105 & 0.8 & day, normal & coarse sand \\ \cline{3-10} 
&  & \multirow{5}{*}{wanke}
 & slow & forward & 4 & 142 & 0.1  & day, weak & fine sand \\
 &  &  & slow & backaward & 5 & 137 & 0.3 & day, normal & grassy, fine sand \\
 &  &  & slow & backaward & 6 & 137 & 0.3 & day, normal & fine sand \\ 
 &  &  & medium & backaward & 7 & 145 & 0.4 & day, normal & fine sand \\
 &  &  & fast & forward, planar & 8 & 176 & 0.8 & day, strong & fine sand \\ \cline{2-10}
&  \multirow{6}{*}
{\begin{tabular}[c]{@{}c@{}}
Quadruped \\ Robot 
\end{tabular}}  
& \multirow{2}{*}{shanhaili}
 & slow & forward, jerky & 9 & 153 & 0.4 & day, weak & coarse sand \\ 
 &  &  & fast & forward, jerky & 10 & 74 & 0.5 & day, weak & coarse sand \\ \cline{3-10} 
&  & \multirow{4}{*}{wanke} 
 & slow & forward, jerky & 11 & 144 & 0.3 & day, strong & fine sand \\   
 &  &  & fast & forward, jerky & 12 & 148 & 0.7 & day, strong & fine sand \\
 &  &  & fast & backward, jerky & 13 & 155 & 0.3-0.7 & day, normal & fine sand \\
 &  &  & fast & multiple loops, jerky & 14 & 194 & 0.5 & day, normal & fine sand \\ \midrule[0.04cm]

\multirow{10}{*}
{\begin{tabular}[c]{@{}c@{}}
\textbf{TAIL-Plus}
\end{tabular}} 
&  \multirow{6}{*}
{\begin{tabular}[c]{@{}c@{}}
Wheeled \\ Robot
\end{tabular}} 
&  \multirow{3}{*}{shanhaili} 
 & slow & multiple loops & 1 & 407 & 0.3 & day, normal & coarse sand \\
 &  &  & slow & forward & 2 & 197 & 0.4 & night & coarse sand \\ 
 &  &  & medium & forward & 3 & 280 & 0.4 & night & coarse sand \\ \cline{3-10} 
&  & \multirow{3}{*}{wanke}
 & medium & multiple loops & 4 & 216 & 0.5 & day, normal & fine sand \\
 &  &  & medium & forward & 5 & 150 & 0.5 & night & fine sand \\
 &  &  & fast & forward & 6 & 169 & 1.2 & night & fine sand \\ 
\cline{2-10}
&  \multirow{4}{*}
{\begin{tabular}[c]{@{}c@{}}
Quadruped \\ Robot 
\end{tabular}}  
& \multirow{4}{*}{wanke} 
 & slow & backward, jerky & 7 & 134 & 0.4 & night & fine sand \\ 
 &  &  & fast & backward, jerky & 8 & 97 & 0.6 & night & fine sand \\
 &  &  & fast & backward, jerky & 9 & 113 & 0.4 & night & fine sand \\
 &  &  & fast & multiple loops, jerky & 10 & 175 & 0.6 & day, weak & fine sand \\
\bottomrule[0.04cm]
\multicolumn{10}{l}{
Seq. means sequences.
$||\overline{\bm{v}}||$ means the average commanded linear velocity.
T means total time.
}\\
\end{tabular}}}
\label{tab:dataset_summary}
\vspace{-0.3cm}
\end{table*}

The calibration chain is shown in Fig. \ref{fig:calib}. We use Allan Variance ROS\footnote{\url{https://github.com/ori-drs/allan_variance_ros}} for IMU intrinsic calibration and MATLAB Camera Calibrator\footnote{\url{https://www.mathworks.com/help/vision/ref/cameracalibrator-app.html}} for color cameras' intrinsics calibration. 
The IMU-Camera extrinsics are calibrated using Kalibr toolbox \cite{kalibr} and LiDAR-Camera extrinsics are calibrated by MATLAB LiDAR Camera Calibrator\footnote{\url{https://www.mathworks.com/help/lidar/ug/get-started-lidar-camera-calibrator.html}} shown in Fig. \ref{fig:calib-cll}. 
The stereo parameters are obtained by Stereo Camera Calibrator\footnote{\url{https://www.mathworks.com/help/vision/ug/using-the-stereo-camera-calibrator-app.html}}.

\section{Data collection}
\label{sec:dataset}

As mentioned before, we choose beaches as analog environments for their unstructured, deformable sandy terrains. The data were collected on beaches at the Double-Moon Bay, Guangdong Province. The east coast with coarse-sand beach is called \texttt{shanhaili}, and the west coast with fine-sand beach is called \texttt{wanke}. As shown in Fig. \ref{fig:sample_data}, we conducted field experiments with various traverses on these two beaches. The 6-DoF reference ground-truth poses are provided by IMU-integrated RTK-GPS measurements.

In TAIL-Plus dataset, we release 10 sequences covering different robot locomotion (wheeled/quadruped robot), illumination change (day/night) and various surface characteristics (coarse/fine sand). Tab. \ref{tab:dataset_summary} briefly shows the features of the TAIL-Plus sequences as well as the original TAIL sequences. 
7 sequences of TAIL-Plus were recorded during night, and other 3 sequences with multiple loop were recorded during daylight.
These sequences provide challenging scenes for testing multi-sensor fusion algorithms for field robots in planetary surface analog environments with sandy terrains.

\section{Conclusion}
\label{sec.conclusion}
This paper release TAIL-Plus dataset, an update and extension to \textbf{TAIL} (\textbf{T}errain-\textbf{A}ware Mult\textbf{I-}Moda\textbf{L}) dataset, for robot localization and mapping in unstructured, deformable granular terrains, especially in planetary surface exploration missions. Utilizing both wheeled and legged robot platforms, we provide time-synchronized, spatially-calibrated multi-sensor data with a wide range of conditions, including different types of locomotion, multi-loops, day-night illumination change and surface terrain change.
Our datasets can be used for testing various kinds of multi-sensor SLAM algorithms, like LiDAR-inertial, visual-inertial and LiDAR-visual-inertial fusions.

As for the future work, we would like to update our sensor suite to equip more types of sensors, such as event cameras, infrared thermal cameras and solid-state LiDARs, and utilize more types of robots like Unmanned Aerial Vehicles (UAVs). Besides, we will extend the datasets to cover more challenging environments and corner cases.

\section*{ACKNOWLEDGMENT}
We appreciate the contributors of the open-source tools of FusionPortable \cite{jiao2022fusionportable} and VECtor \cite{gao2022vector}. We would also like to thank Yuntian Zhao for his help in experiments.
 
\bibliographystyle{IEEEtran}
\bibliography{refs}

\end{document}